\documentclass[10pt,twocolumn,letterpaper]{article}

\usepackage{cvpr}
\usepackage{times}
\usepackage{epsfig}
\usepackage{graphicx}
\usepackage{amsmath}
\usepackage{amssymb}
\usepackage{subfigure}
\usepackage[ruled]{algorithm}
\usepackage{algpseudocode}
\usepackage{bbm}
\usepackage{verbatim}
\usepackage[pagebackref=false,breaklinks=true,letterpaper=true,colorlinks,bookmarks=false]{hyperref}

\cvprfinalcopy % *** Uncomment this line for the final submission

\newcommand{\R}{\mathbb{R}}

\usepackage{array}
\newcommand*\rot{\rotatebox{90}}
\newcolumntype{C}[1]{>{\centering\let\newline\\\arraybackslash\hspace{0pt}}m{#1}}
\newcolumntype{L}[1]{>{\raggedright\let\newline\\\arraybackslash\hspace{0pt}}m{#1}}
\usepackage{makecell}
\usepackage{multirow}
\usepackage{float}
\newcounter{savecntr}% Save footnote counter
\newcounter{restorecntr}% Restore footnote counter

\def\cvprPaperID{1968} % *** Enter the CVPR Paper ID here

\ifcvprfinal\fi
\begin{document}

%%%%%%%%% TITLE
\title{Phase Consistent Ecological Domain Adaptation}

\author{Yanchao Yang\setcounter{savecntr}{\value{footnote}}\thanks{These two authors contributed equally. Please send correspondence to yanchao.yang@cs.ucla.edu and dong.lao@kaust.edu.sa.}\\
UCLA Vision Lab\\
%{\tt\small yanchao.yang@cs.ucla.edu}
\and
Dong Lao\setcounter{restorecntr}{\value{footnote}}\setcounter{footnote}{\value{savecntr}}\footnotemark\setcounter{footnote}{\value{restorecntr}}\\
KAUST\\
%{\tt\small dong.lao@kaust.edu.sa}
\and
Ganesh Sundaramoorthi\\
KAUST \& UTRC\\
%{\tt\small ganesh.sun@gmail.com}
\and
Stefano Soatto\\
UCLA Vision Lab\\
%{\tt\small soatto@cs.ucla.edu}
}

\maketitle
\thispagestyle{empty}

\begin{abstract}
We introduce two criteria to regularize the optimization involved in learning a classifier in a domain where no annotated data are available, leveraging annotated data in a different domain, a problem known as unsupervised domain adaptation. 
We focus on the task of semantic segmentation, where annotated synthetic data are aplenty, but annotating real data is laborious. 
The first criterion, inspired by visual psychophysics, is that the map between the two image domains be phase-preserving. 
This restricts the set of possible learned maps, while enabling enough flexibility to transfer semantic information. 
The second criterion aims to leverage ecological statistics, or regularities in the scene which are manifest in any image of it, regardless of the characteristics of the illuminant or the imaging sensor.
It is implemented using a deep neural network that scores the likelihood of each possible segmentation given a single un-annotated image.
Incorporating these two priors in a standard domain adaptation framework improves performance across the board in the most common unsupervised domain adaptation benchmarks for semantic segmentation.\footnote{Code available at: https://github.com/donglao/PCEDA}
\end{abstract}

\vspace{-0.1cm}
\section{Introduction}

Unsupervised domain adaptation (UDA) aims to leverage an annotated ``source'' dataset in designing learning schemes for  a ``target'' dataset for which no ground-truth is available. 
This problem arises when annotations are easy to obtain in one domain (\eg, synthetic images) but expensive in another (\eg, real images), and is exacerbated in tasks where the annotation is laborious, as in semantic segmentation where each pixel in an image is assigned one of $K$ labels.
If the two datasets are sampled from the same distribution, this is a standard semi-supervised learning problem. 
The twist in UDA is that the distributions from which source and target data are drawn differ enough that a model trained on the former performs poorly, out-of-the-box, on the latter.

\begin{figure}[!t]
\begin{center}
  \includegraphics[width=0.4\textwidth]{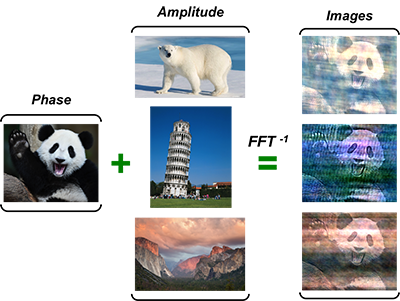}
\end{center}
\vspace{-0.2cm}
\caption{\sl\small {\bf Semantic content is mainly carried by the phase component of the Fourier Transform.} Replacing the amplitude component of the image depicting a panda by the amplitude from other images within a wide range, will not prevent us from recognizing a panda in the images generated by the inverse Fourier Transform.}
\vspace{-0.2cm}
\label{fig:phase-constraint}
\end{figure}

Typical domain adaptation work employing deep neural networks (DNNs) proceeds by either learning a map that aligns the source and target (marginal) distributions, or by training a backbone to be insensitive to the domain change through an auxiliary discrimination loss for the domain variable. 
Either way, these approaches operate on the marginal distributions, since the labels are not available in the target domain. 
However, the marginals could be perfectly aligned, yet the labels could be scrambled: Trees in one domain could map to houses in another, and vice-versa. 
Since we want to transfer information about the classes, ideally we would want to align the class-conditional distributions, which we do not have.
Recent improvements in UDA, for instance cycle-consistency, only enforce the invertibility of the map, but not preservation of semantic information such as the class identity, see Fig. \ref{fig:phase_result}. Since the problem is ill-posed, constraints or prior have to be enforced in UDA. 

We introduce two priors or constraints, one on the map between the domains, the other on the classifier in the target domain, both unknown at the outset. 

For the map between domains, we know from visual psychophysics that semantic information in images tends to be associated with the phase of its Fourier transform. Changes in the amplitude of the Fourier transform can significantly alter the appearance but not the interpretation of the image. This suggests placing an incentive for the transformation between domains to be phase-preserving. Indeed, we show from ablation studies that imposing phase consistency improves the performance of current UDA models.

For the classifier in the target domain, even in the absence of annotations, a target image informs the set of possible hypotheses (segmentations), due to the statistical regularities of natural scenes (ecological statistics, \cite{brunswik1953ecological,elder2002ecological}). 
Semantic segments are unlikely to straddle many boundaries in the image plane, and their shape is unlikely to be highly irregular due to the regularity of the shape of objects in the scene. 
Such generic priors, informed by each single un-annotated images, could be learned from other (annotated) images and transfer across image domains, since they arise from properties of the scene they portray. 
We use a Conditional Prior Network \cite{yang2018conditional} to learn a data-dependent prior on segmentations that can be imposed in an end-to-end framework when learning a classifier in the target domain in UDA.

These two priors yield improvement in UDA benchmarks. We conduct ablation studies to quantify the effect of each prior on the overall performance of learned classifiers (segmentation networks).

In the next section, we describe current approaches to UDA and then describe our method, which is summarized in Sect. \ref{sec:final}, before testing it empirically in Sect. \ref{sec:experiments}.

\subsection{Related Work}

Early works on UDA mainly focus on image classification \cite{glorot2011domain,fernando2013unsupervised,baktashmotlagh2013unsupervised}, by minimizing a discrepancy measure between two domains \cite{geng2011daml}. Recent methods apply adversarial learning \cite{ganin2014unsupervised,tzeng2017adversarial} for classification, by instantiating a discriminator that encourages the alignment in feature space \cite{sener2016learning,kumar2018co,shu2018dirt}. Unfortunately, none of these methods achieves the same success on semantic segmentation tasks.

Recent progress in image-to-image transformation techniques \cite{CycleGAN2017,liu2017unsupervised} aligns domains in image space, with some benefit to semantic segmentation \cite{hoffman2016fcns,hoffman2017cycada}. \cite{hoffman2016fcns} is the first UDA semantic segmentation method utilizing both global and categorical adaptation techniques. CyCADA \cite{hoffman2017cycada} adapts representations in both image and feature space while enforcing cycle-consistency to regularize the image transformation network. \cite{sankaranarayanan2018learning} also applies image alignment by projecting the learned intermediate features into the image space. \cite{zhang2017curriculum} proposes curriculum learning to gradually minimize the domain gap using anchor points. \cite{wu2018dcan} reduces domain shift at both image and feature levels by aligning statistics in each channel of CNN feature maps in order to preserve spatial structures. \cite{gong2019dlow} generates a sequence of intermediate shifted domains from source to target to further improve the transferability by providing multi-style translations. \cite{luo2019taking} introduces a category-level adversarial network to prevent the degeneration of well-aligned categories during global alignment. \cite{hong2018conditional} conditions on both source images and random noise to produce samples that appear similar to the target. Despite the difficulty in training the domain discriminators, generally, the alignment criteria provided by the domain discriminators do not guarantee consistency of the semantic content between the original and transformed images. In addition to cycle-consistency, \cite{li2019bidirectional,Chen_2019_CVPR} propose using the segmentation network on the target domain to encourage better semantic consistency. However, this will make the performance depend highly on the employed surrogate network.

In psychophysics, \cite{piotrowski1982demonstration} demonstrates that certain phase modifications can hinder or prevent the recognition of visual scenes. \cite{oppenheim1981importance} shows that many important features of a signal can be preserved by the phase component of the Fourier Transform, and under some conditions a signal can be completely reconstructed with only the phase. Moreover, \cite{hansen2007structural} shows psychophysically that the Fourier phase spectrum plays a critical role in human vision. 
The concurrent work \cite{yang2020fda} shows that swapping the amplitude component of an image with one from the other domain preserves the semantic content while aligning the two domains.
With all these observations, we propose to use phase to provide an effective semantic consistency constraint that does not depend on any surrogate networks.

Besides discriminators applied to the image or in feature space, \cite{tsai2018learning, tsai2019domain} find that adaptation on the structured output space is also beneficial for semantic segmentation. \cite{chen2018road} proposes spatially-aware adaptation along with target guided distillation using activation supervision with a pretrained classification network. Further, \cite{chen2019learning} proposes a geometrically guided adaptation aided with depth in a multi-task learning framework. \cite{chang2019all} extracts the domain invariant structure from the image to disentangle images into domain invariant structure and domain-specific variations. \cite{zou2018unsupervised} performs iterative class-balanced self-training as well as refinement of the generated pseudo-labels using a spatial prior. A similar strategy is also applied in \cite{li2019bidirectional,tsai2019domain}. \cite{vu2019advent} approaches UDA for semantic segmentation by entropy minimization of the pixel-wise predictions. An adversarial loss on the entropy map is also used to introduce regularity in the output space. However, none of them explicitly models the scene compatibility that regularizes the training of the target domain segmentation network.

\vspace{-0.1cm}
\section{Method}
\vspace{-0.1cm}

\def\figd{figures/phase}
\def\fWidD{0.15\textwidth}
\begin{figure}[!ht]
\centering
{\footnotesize
\begin{tabular}{c@{\hspace{0.01in}}c@{\hspace{0.01in}}c}\\
Image&Cycle Consistency&Phase Consistency\\
\includegraphics[width=\fWidD]{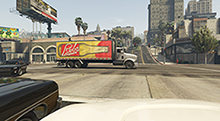}&\includegraphics[width=\fWidD]{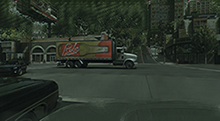}&\includegraphics[width=\fWidD]{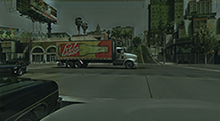}
\vspace{-0.07cm}\\
\includegraphics[width=\fWidD]{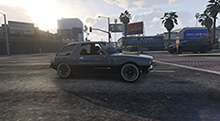}&\includegraphics[width=\fWidD]{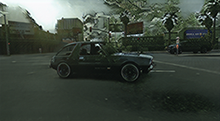}&\includegraphics[width=\fWidD]{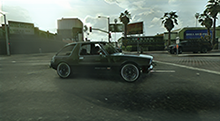}
\vspace{-0.07cm}\\
\includegraphics[width=\fWidD]{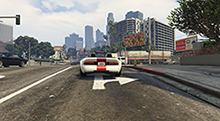}&\includegraphics[width=\fWidD]{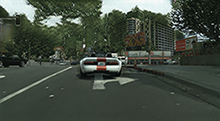}&\includegraphics[width=\fWidD]{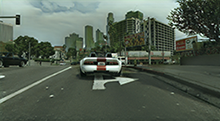}
\vspace{-0.07cm}\\
\end{tabular}
}
\caption{\sl\small {\bf Cycle consistency does not guarantee semantic consistency, but phase does.} Note the sky is transformed to trees (1st row), the cloud is transformed to a mountain (2nd row), and the buildings are also transformed to trees (3rd row) even if cycle consistency is enforced. Phase consistency enforces the semantic information to be preserved and provides enough flexibility to align the two domains.}
\vspace{-0.3cm}
\label{fig:phase_result}
\end{figure}

We first describe general image translation for unsupervised domain adaptation (UDA) and how it is used in semantic segmentation. We point to some drawbacks as inspiration for the two complementary constraints, which we introduce in Sect. \ref{sec:phase-constraint} and \ref{sect:cpn-on-phi}, and incorporate into a model of UDA for semantic segmentation in Sect. \ref{sec:final}, which we validate empirically in Sect. \ref{sec:experiments}.

\subsection{Preliminaries: Image Translation for UDA}

We consider two probabilities, a source $P^s$ and a target $P^\tau$, which are generally different (covariate shift), as measured by the Kullbach-Liebler divergence $KL(P^s || P^\tau)$. In UDA, we are given ground-truth annotation in the source domain only. So, if $x \in \R^{H\times W \times 3}$ are color images, and $y \in [1, \dots, K]^{H\times W}$ are segmentation masks where each pixel has an associated label, we have images and segmentations in the source domain, $D^s = \{ (x^s_i, y^s_i) \sim P^s(x,y) \}_{i=1}^{N_s} $ but only images in the target domain, $\{x_i^\tau \sim P^\tau(x)\}_{i=1}^{N_\tau}$. 
The goal of UDA for semantic segmentation is to train a model $\phi^\tau$, for instance a deep neural network (DNN), that maps target images to estimated segmentations, $x^\tau \mapsto \hat y^\tau = \arg\max_y \phi^\tau(x^\tau)_y$, leveraging source domain annotations. 
Because of the covariate shift, simply applying to the target data a model trained on the source generally yields disappointing results. As observed in \cite{ben2010theory}, the upper bound on the target domain risk can be minimized by reducing the gap between two distributions. 
Any invertible map $T$ between samples in the source and target domains, for instance $x^s \mapsto T(x^s)$ induces a (pushforward) map between their distributions $P^s \mapsto T_* P^s$ where $T_*P^s(x^\tau) = P^s(T^{-1}(x^\tau))$. The map can be implemented by a ``transformer'' network, and the target domain risk is minimized by the cross-entropy loss, whose empirical approximation is:
\begin{equation}
    \mathcal{L}_{ce}(\phi^\tau, T; D^s) = \sum_{(x_i, y_i) \in D^s} - \log [\phi^\tau(T(x_i))]_{y_i} 
\label{eq:segmentation-loss}
\end{equation}
where $T$ maps data sampled from the source distribution to the target domain. The gap is measured by $KL(P^\tau || T_*P^s)$, and can be minimized by (adversarially) maximizing the domain confusion, as measured by a domain discriminator $\theta$ that maps each image into the probability of it coming from the source or target domains:
\begin{equation}
    \mathcal{L}_D(\theta, T; x^s_i) = -\log[ \theta(T(x^s_i)) ].
\label{eq:discriminator-loss-T}
\end{equation}
Ideally, $\theta$ returns $1$ for images drawn from the target $P^\tau$, and $0$ otherwise.

\subsection{Limitations and Challenges}

Ideally, jointly minimizing the two previous equations would yield a segmentation model that operates in the target domain, producing estimated segmentations $y^\tau = \phi(x^\tau)$.
Unfortunately, a transformation network $T$ trained by minimizing Eq. \eqref{eq:discriminator-loss-T} does not yield a good target domain classifier, as $T$ is only asked by Eq. \eqref{eq:discriminator-loss-T} to match the marginals, which it could do while scrambling all labels (images of class $i$ in the source can be mapped to images of class $j$ in the target). In other words, the transformation network can match the image statistics, but there is nothing that encourages it to match semantics. Cycle-consistency \cite{CycleGAN2017,hoffman2017cycada} does not address this issue, as it only enforces the invertibility of $T$:
\begin{equation}
    \mathcal{L}_{cyc}(T, T^{-1}; x^s_i) = \| x^s_i - T^{-1}\circ T(x^s_i) \|_{1}.
\label{eq:cycle-consistency-T}
\end{equation}
Even after imposing this constraint, buildings in the source domain could be mapped to trees in the target domain, and vice-versa (Fig. \ref{fig:phase_result}). Ideally, if $\phi^s$ is a model trained on the source, and $\phi^\tau$ the one operating on the target, we would like:
\begin{equation}
    \phi^s(x^s_i) = {\phi^\tau}( T(x^s_i)), \forall i.
    \label{eq:ideal-semantic-consistency}
\end{equation}
Unfortunately, training ${\phi^\tau}$ would require ground-truth in the target domain, which is unknown. We could use $\phi^s$ as a surrogate, apply $\phi^s$ on the target domain, and penalize the discrepancy between the two sides in Eq. \eqref{eq:ideal-semantic-consistency} with respect to the unknowns. Absent any regularization, this yields the trivial result where $T(x) = x$ and $\phi^\tau = \phi^s$. While Eq. \eqref{eq:ideal-semantic-consistency} is useless in providing information on $T$ {\em and} $\phi^\tau$, it can be seen as a vehicle to transfer prior information {\em from one} (\eg, $T$) {\em onto the other} (\eg., $\phi^\tau$). In the next two sections we discuss additional constraints and priors that can be imposed on $T$ (Sect. \ref{sec:phase-constraint}) and $\phi^\tau$ (Sect. \ref{sect:cpn-on-phi}) that make the above constraint non-trivial, and usable in the context of UDA.

\subsection{Phase Consistency}
\label{sec:phase-constraint}

It is well known in perceptual psychology that manipulating the spectrum of an image can lead to different effects: Changes in the amplitude of the Fourier transform alters the image but does not affect its interpretation, whereas altering the phase produces uninterpretable images \cite{kermisch1970image,piotrowski1982demonstration,oppenheim1981importance,hansen2007structural}. This is illustrated in Fig. \ref{fig:phase-constraint}, where the amplitude of the Fourier transform of an image of a panda is replaced with the amplitude from an image of a bear, a tourist landmark and a landscape, yet the reconstructed images portray a panda. In other words, it appears that semantic information is included in the phase, not the amplitude, of the spectrum. This motivates us to hypothesize that the transformation $T$ should be phase-preserving.

To this end, let $\mathcal{F}: \R^{H\times W} \rightarrow \R^{H\times W \times 2}$ be the Fourier Transform. Phase consistency, for a transformation $T$, for a single channel image $x$, is obtained by minimizing:
\begin{equation}
    \mathcal{L}_{ph}(T; x) = - \sum_{j} \dfrac{\langle\mathcal{F}(x)_j, \mathcal{F}(T(x))_j\rangle}{\|\mathcal{F}(x)_j\|_{2}\cdot\|\mathcal{F}(T(x))_j\|_{2}}
    \label{eq:phase-constraint}
\end{equation}
where $\langle , \rangle$ is the dot-product, and $\| \cdot \|_{2}$ is the $L_2$ norm. Note that Eq. \eqref{eq:phase-constraint} is the negative cosine of the difference between the original and the transformed phases, thus, by minimizing Eq. \eqref{eq:phase-constraint} we can directly minimize their difference and increase phase consistency. We demonstrate the effectiveness of phase consistency in the ablation studies in Sect. \ref{sec:experiments}.

\subsection{Prior on Scene Compatibility} \label{sect:cpn-on-phi}

While target images have no ground-truth labeling, not all semantic segmentations are equally likely at the outset.
Given an unlabeled image, we may not know what classes $\{1, \dots, K\}$ may interest a user, but we do know that objects in the scene have certain regularities, so it is unlikely that photometrically homogeneous regions are segmented into many pieces, or that a class segment straddles many image boundaries. 
It is also unlikely that the segmented map is highly irregular. These characteristics inform the probability of a segmentation given the image in the target domain, $Q(\phi(x)| x)$. $Q$ can be thought of as a function that scores each hypothesis $\phi(x)$ based on the plausibility of the resulting segmentation given the input image $x$. 
The function can be learned using images for which the ground-truth segmentation is given, for instance the source dataset $D^s$, and then used at inference time as a scoring function. Such a scoring function can be implemented by a Conditional Prior Network (CPN) \cite{yang2018conditional}. However, note that $D^s=\{ (x^s_i,y^s_i) \}$ is sampled from $P^s(x,y)$. Simply training a CPN with $D^s$ will make $Q(y| x)$ approximate $P^s(y|x)$\footnote{We abuse the notation and use $y$ to indicate both the class and the soft-max (log-likelihood) vector that approximates its indicator function.}, making the exercise moot. The CPN would capture both the domain-related unary prediction term and the domain irrelevant pairwise term that depends on the image structure. To make this point explicit, we can decompose $P^s(y|x)$ as follows:

\begin{equation}
    P^s(y|x) \approx \prod_{j}P^s(y_j|x) \prod_{m < n} P(y_m= y_n|x)
\label{eq:ps-dcomp}
\end{equation}
where we omit higher-order terms for simplicity. The unary terms $P^s(y_j|x)$ measure the likelihood of the semantic label of a single pixel given the image; \eg, pixels in a white region indicate sky in the source domain, which depends highly on the domains. The pairwise terms $P(y_m=y_n|x)$ measure the labeling compatibility between pixels, which would depend much less on the domain; \eg, pixels in a white region may not be sky in the target domain, but they should be labeled the same. Absent at least binary terms, the unary terms would lead to overfitting the source domain. To prevent this, we randomly permute the labels in $y^s$ according to a uniform distribution:
\begin{equation}
    y^s\mid_{y^s=i} \ = \mathrm{PM}^{K}(i)
    \label{eq:permutate}
\end{equation}

\begin{figure}[!t]
\begin{center}
  \includegraphics[width=0.3\textwidth]{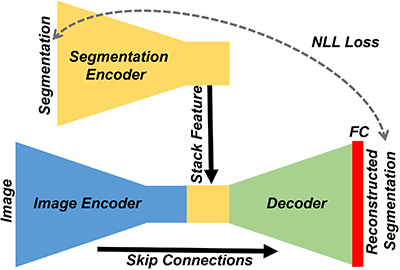}
\end{center}
\vspace{-0.1cm}
\caption{\sl\small {\bf CPN Architecture.} To reconstruct the segmentation map that is encoded into a narrow bottleneck, the decoder needs to leverage structural information from the image. The CPN reconstructs better the prediction $\phi(x)$ with higher compatibility with $x$. Labels are randomly permuted during training to avoid overfitting to the domain dependent unary prediction terms in Eq. \eqref{eq:ps-dcomp}.}
\vspace{-0.05cm}
\label{fig:CPN}
\end{figure}

\noindent where $\mathrm{PM}^K$ is a random permutation of the class ID's for $K$ classes, and we denote the permuted semantic segmentation masks as $\hat{y^s}$, which scales the original dataset up in size by a factor of $K!$. We denote the new source dataset with permuted ground-truth masks as $\hat{D^s}$, which will render the conditional distribution invariant to the domain-dependent unaries, \ie:
\begin{equation}
    \hat{P^s}(y|x) \approx \prod_{m<n}P(y_m=y_n|x)
\label{eq:ps-permute}
\end{equation}
Note, $\hat{P^s}(y|x)$ only evaluates the compatibility based on the segmentation layout but not the semantic meanings. Thus, we train a CPN $Q$ using the following training loss \cite{yang2018conditional} with an information capacity constraint:
\begin{equation}
    \min_{Q} \mathbb{E}_{x} {KL}( \hat{P^s}(y|x), Q(y|x)) + \beta \mathbb{I}( y, Q^e(y))
    \label{eq:loss-cpn}
\end{equation}
where $\mathbb{I}$ denotes the mutual information between $y$ and its CPN encoding $Q^e(y)$. Then, we obtain a compatibility function
\begin{equation}
    Q(y|x) \approx \prod_{m<n}P(y_m=y_n|x)
\end{equation}

The proposed CPN architecture is illustrated in Fig. \ref{fig:CPN} and the training details, including the encoding metric, is described in Sect. \ref{sec:experiments}. We now summarize the overall training loss, that exploits regularities implied by each constraint.

\begin{table*}[!ht]
\footnotesize
\begin{tabular}{|c|c|c|p{0.3cm}<{\centering}p{0.3cm}<{\centering}p{0.3cm}<{\centering}p{0.3cm}<{\centering}p{0.3cm}<{\centering}p{0.3cm}<{\centering}p{0.3cm}<{\centering}p{0.3cm}<{\centering}p{0.3cm}<{\centering}p{0.3cm}<{\centering}p{0.3cm}<{\centering}p{0.3cm}<{\centering}p{0.3cm}<{\centering}p{0.3cm}<{\centering}p{0.3cm}<{\centering}p{0.3cm}<{\centering}p{0.2cm}<{\centering}p{0.3cm}<{\centering}p{0.4cm}<{\centering}|c|}
\hline
\rot{Architecture}&SSL&CPN&\rot{road}&\rot{sidewalk}&\rot{building}&\rot{wall}&\rot{fence}&\rot{pole}&\rot{light}&\rot{sign}&\rot{vegetation}&\rot{terrain}&\rot{sky}&\rot{person}&\rot{rider}&\rot{car}&\rot{truck}&\rot{bus}&\rot{train}&\rot{motorcycle}&\rot{bicycle}&mIoU
\\
\hline
A&&&88.2&41.3&\bf{83.2}&28.8&21.9&\bf{31.7}&\bf{35.2}&28.2&\bf{83.0}&26.2&\bf{83.2}&57.6&27.0&77.1&27.5&34.6&2.5&28.3&\bf{36.1}&44.3\\
A&&\checkmark&\bf{91.4}&\bf{47.2}&82.9&\bf{29.2}&\bf{22.9}&31.4&33.3&\bf{30.2}&80.8&\bf{27.8}&81.3&\bf{59.1}&\bf{27.7}&\bf{84.4}&\bf{31.5}&\bf{40.9}&\bf{3.2}&\bf{30.2}&24.5&\bf{45.3}\\
\cline{2-23}
A&\checkmark&&91.2&46.1&83.9&31.6&20.6&29.9&36.4&31.9&\bf{85.0}&39.7&\bf{84.7}&57.5&29.6&83.1&38.8&\bf{46.9}&\bf{2.5}&27.5&38.2&47.6\\
A&\checkmark&\checkmark&\bf{91.3}&\bf{48.2}&\bf{85.0}&\bf{39.4}&\bf{26.1}&\bf{32.4}&\bf{37.4}&\bf{40.7}&84.9&\bf{41.9}&83.0&\bf{59.8}&\bf{30.2}&\bf{83.6}&\bf{40.0}&46.1&0.1&\bf{31.7}&\bf{43.3}&\bf{49.7}\\
\hline
B&&&\bf{86.4}&39.5&79.2&27.4&\bf{24.3}&\bf{23.4}&29.0&18.0&80.5&33.2&70.1&47.2&18.1&75.4&20.6&23.3&0.0&\bf{16.1}&5.4&37.7\\
B&&\checkmark&86.0&\bf{39.9}&\bf{80.6}&\bf{32.3}&21.9&21.6&\bf{29.5}&\bf{23.9}&\bf{83.1}&\bf{37.5}&\bf{75.9}&\bf{53.2}&\bf{24.4}&\bf{79.3}&\bf{22.8}&\bf{32.4}&\bf{0.9}&13.9&\bf{18.9}&\bf{40.9}\\
\cline{2-23}
B&\checkmark&&89.2&40.9&\bf{81.2}&29.1&19.2&14.2&29.0&19.6&\bf{83.7}&35.9&\bf{80.7}&\bf{54.7}&23.3&\bf{82.7}&\bf{25.8}&28.0&\bf{2.3}&\bf{25.7}&19.9&41.3\\
B&\checkmark&\checkmark&\bf{90.1}&\bf{44.7}&81.0&\bf{29.3}&\bf{26.4}&\bf{20.9}&\bf{33.7}&\bf{34.3}&83.4&\bf{37.4}&71.2&54.0&\bf{27.4}&79.9&23.7&\bf{39.6}&1.1&18.5&\bf{22.6}&\bf{43.1}\\
\hline
C&&&79.1&33.1&77.9&\bf{23.4}&\bf{17.3}&32.1&33.3&31.8&81.5&26.7&69.0&\bf{62.8}&\bf{14.7}&74.5&\bf{20.9}&\bf{25.6}&6.9&\bf{18.8}&\bf{20.4}&39.5\\
C&&\checkmark&\bf{89.1}&\bf{41.4}&\bf{81.2}&22.2&15.3&\bf{34.0}&\bf{35.0}&\bf{37.1}&\bf{84.8}&\bf{32.1}&\bf{76.2}&61.7&12.5&\bf{82.1}&20.8&25.2&\bf{7.3}&15.6&18.9&\bf{41.7}\\
\hline
\end{tabular}
\vspace{0.15cm}
\caption{\sl\small {\bf The learned Scene Compatibility improves segmentation accuracy.} Training the segmentation model with the learned scene compatibility $Q$ improves segmentation accuracy under all experimental settings, with different network backbones: A: ResNet-101, B: VGG-16, C: DRN-26. SSL: Self-supervised Learning. Note that whenever $Q$ is added in the training loss while the other terms are fixed, the overall semantic segmentation performance gets improved.}
\vspace{-0.1cm}
\label{tab:cpn_result}
\end{table*}

\subsection{Overall Training Loss}
\label{sec:final}

Combining the adversarial losses and our novel constraints for both phase consistency and scene compatibility, we have the overall training loss for the proposed domain adaptation method for training the image transformation networks $T, T^{-1}$ and the target domain segmentation network $\phi^\tau$:
%\begin{multline}
%    \mathcal{L}(\phi^\tau, T^{s\to\tau}, T^{\tau\to s}; \theta^s, \theta^\tau, x^s_i, y^s_i, x^\tau_i) = \\
 %   \lambda_D( \mathcal{L}_D(\theta^\tau, T^{s\to\tau}; x^s_i) + \mathcal{L}_D(\theta^s, T^{\tau\to s}; x^\tau_i) )\\
%    + \lambda_{cyc}( \mathcal{L}_{cyc}(T^{s\to\tau}, T^{\tau\to s}; x^s_i) + \mathcal{L}_{cyc}(T^{\tau\to s}, T^{s\to\tau}; x^\tau_i) )\\
%    + \lambda_{ph}( \mathcal{L}_{ph}(T^{s\to\tau}; x^s_i) + \mathcal{L}_{ph}(T^{\tau\to s}; x^\tau_i) ) \\
%    \mathcal{L}_{ce}(\phi^\tau, T^{s\to \tau}; x^s_i, y^s_i) - \lambda_{cpn}\log [Q_{cpn}(\phi^\tau(x^\tau_i)|x^\tau_i)]
%\label{eq:final-training-loss}
%\end{multline}
\begin{multline}
    \mathcal{L}(\phi^\tau, T, T^{-1}; \theta^s, \theta^\tau, x^s_i, y^s_i, x^\tau_i) = \\
    \lambda_D( \mathcal{L}_D(\theta^\tau, T; x^s_i) + \mathcal{L}_D(\theta^s, T^{-1}; x^\tau_i) )\\
    + \lambda_{cyc}( \mathcal{L}_{cyc}(T, T^{-1}; x^s_i) + \mathcal{L}_{cyc}(T^{-1}, T; x^\tau_i) )\\
    + \lambda_{ph}( \mathcal{L}_{ph}(T; x^s_i) + \mathcal{L}_{ph}(T^{-1}; x^\tau_i) ) \\
    \mathcal{L}_{ce}(\phi^\tau, T; x^s_i, y^s_i) - \lambda_{cpn}\log [Q(\phi^\tau(x^\tau_i)|x^\tau_i)]
\label{eq:final-training-loss}
\end{multline}
with $\lambda$'s the corresponding weights on each term (hyperparameters), whose values will be reported in Sect.\ref{sec:experiments}. Note when training $\phi^\tau$ using Eq. \eqref{eq:final-training-loss}, we do not permute its output to evaluate the scene compatibility term. And the scene compatibility $Q$ is fixed after it is trained using Eq. \eqref{eq:loss-cpn}. We follow the standard procedure in \cite{hoffman2017cycada,li2019bidirectional} to train the domain discriminators.

\section{Experiments}
\label{sec:experiments}

We evaluate the proposed UDA method on synthetic-to-real semantic segmentation tasks, where the source images (GTA5 \cite{Cordts2016Cityscapes} and Synthia \cite{RosCVPR16synthia}) and corresponding annotations are generated using graphics engines, and the adapted segmentation models are tested on real-world images. We use average intersection-over-union score (mIoU) across semantic classes as the evaluation metric in all experiments. Moreover, the frequency weighted IoU (fwIoU), which is the sum of the IoUs of different classes but weighted by how frequent a certain class appears in the dataset, is calculated and compared in the GTA5-to-Cityscapes experiments.

We first describe the data used for training and the implementation details, followed by a comprehensive ablation study demonstrating the effectiveness of each proposed component in our method. Then we show quantitative and qualitative comparisons against the state-of-the-art methods, using networks with different backbones, on the GTA5-to-Cityscapes and Synthia-to-Cityscapes benchmarks.

\subsection{Datasets}

\textbf{Cityscapes} \cite{Cordts2016Cityscapes} is a real-world semantic segmentation dataset containing 2975 street view training images and 500 validation images with original resolution $2048 \times 1024$, which is resized to $1024 \times 512$ for training. The images are collected during the day in multiple European cities and densely annotated. We train the image transformation network and the adapted segmentation network using the training set, and report the result on the validation set.

\textbf{GTA5} \cite{Richter_2016gta} contains 24966 synthesized images from the Grand Theft Auto game with resolution $1914 \times 1052$. It exhibits a wide range of variations including weather and illumination. We resize the images to $1280 \times 720$ and use the 19 compatible classes for the  training and evaluation.

\textbf{Synthia} \cite{RosCVPR16synthia} is a synthetic dataset focusing on driving scenarios rendered from a virtual city. We use the SYNTHIA-RAND-CITYSCAPES subset as source data, which contains 9400 images with the resolution of $1280 \times 760$ for training the 16 common classes with Cityscapes, and we evaluate the trained network using both the 16 classes or a subset of 13 classes following previous works \cite{tsai2018learning,li2019bidirectional,Du_2019_ICCV}.

\subsection{Implementation Details}

\textbf{Image Transformation Network:} We adapt the public CycleGAN \cite{CycleGAN2017} framework, and use the ``cycle$\_$gan'' model therein. We set $\lambda_D=1.0$, $\lambda_{cyc}=10.0$ and $\lambda_{ph}=5.0$ for training the image transformation networks $T, T^{-1}$. Images from source and target domains are resized to $1024 \times 512$ and then cropped to $452\times 452$ before feeding into the network. We set the batch-size to $1.0$ and use ``resnet$\_$9blocks'' as the backbone.

% checking below ....

\begin{table}[!t]
\footnotesize \centering
\begin{tabular}{|c|c|c|c|}
\hline
Method & Surrogate & Output Space & mIoU \\
\hline
\multirow{2}*{CyCADA \cite{hoffman2017cycada}} & \checkmark & & 43.5 \\
    & \checkmark & \checkmark & 43.1 \\
    \hline
\multirow{3}*{AdaptSegNet \cite{tsai2018learning}} & & & 36.6 \\
    & \checkmark & & 39.3 \\
    & \checkmark & \checkmark & 42.4 \\
    \hline
\multirow{3}*{BDL \cite{li2019bidirectional}} & & \checkmark & 41.1 \\
    & \checkmark* & \checkmark & 42.7 \\
    & \checkmark$^\dagger$ & \checkmark & 44.4 \\
    \hline
\multirow{2}*{Ours}& & \checkmark & 44.8 \\
    & & & \bf{45.3} \\
\hline
\end{tabular}
\vspace{0.15cm}
\caption{\sl\small {\bf Phase consistency (ours) achieves better performance.} Note our model trained only with the phase consistency outperforms other methods that utilize a surrogate network to impose semantic consistency (Surrogate), or employ output space regularization (Output Space). * and $^\dagger$: first and second round of improved image transformation using a self-trained surrogate network.}
\vspace{-0.1cm}
\label{tab:phase_result}
\end{table}

\textbf{Conditional Prior Network:} We adopt the standard UNet \cite{RFB15a} architecture, and add the segmentation encoder branch. We instantiate 6 convolutional layers, whose channel numbers are \{16, 32, 64, 128, 256, 256\}, to encode the image. Each of the first five layers is followed by $2 \times 2$ max pooling, similarly, for semantic segmentation maps. Encoded image and segmentation are stacked at the bottleneck, then passed through a 6-layer decoder with channel numbers \{512, 256, 128, 64, 32, 16\}, followed by a fully connected layer for class prediction. Skip connections are enabled between the image encoder and the decoder. The network is trained with batch size four by ADAM optimizer with initial learning rate 1e-4. The learning rate is reduced by a factor of 10 after every 30000 iterations.

During training, the network aims at reconstructing the encoded $\hat{y}^s$, which is the randomly permutated ground-truth segmentation, by utilizing image information, leading to the training loss:
\begin{equation}
    \mathcal{L}_{cpn}(Q; \hat{y}^s, x^s) = \mathrm{NLL}( Q(\hat{y}^s|x^s), \hat{y}^s )
    \label{eq:cpn_loss}
\end{equation}
where $\mathrm{NLL}$ denotes the negative log likelihood loss derived from the KL-divergence term in the CPN training loss Eq. \eqref{eq:loss-cpn}. Lower indicates better scene compatibility \ie higher $Q(y|x)$. Note the information capacity constraint in Eq. \eqref{eq:loss-cpn} is implemented by a structural bottleneck as in \cite{yang2018conditional}.

\textbf{Semantic Segmentation Network:} We experiment with different segmentation network backbones. Due to memory constraint, we choose to train the segmentation network after the transformation networks are trained. We first train from scratch the segmentation network using transformed source images and the corresponding annotations using Eq. \eqref{eq:final-training-loss}. We fix $\lambda_{cpn} = 0.5$ for all the experiments. Finally, we apply the self-supervised training technique as in \cite{li2019bidirectional,tsai2019domain} to further improve the performance on the target domain. We accept the high confidence ($>0.9$) predictions as the pseudo labels. All networks are trained using the ADAM optimizer, with learning rate 2.5e-4, 1e-5, and 1e-4 for ResNet-101, VGG-16, and DRN-26, respectively.

\subsection{Ablation Study}

Here we carry out an ablation study to investigate the effectiveness and robustness of the proposed priors.

\def\figd{figures/cpn}
\def\fWidD{0.12\textwidth}
\begin{figure}[t]
\hspace*{-2.5mm}\centering
{\footnotesize
\begin{tabular}{c@{\hspace{0.01in}}c@{\hspace{0.01in}}c@{\hspace{0.01in}}c}\\
\vspace{-1cm}&&&\\
Image&w/o CPN&w/ CPN&Ground-truth\\
\vspace{-0.07cm}\includegraphics[width=\fWidD]{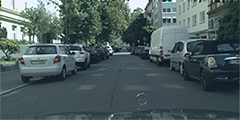}&\includegraphics[width=\fWidD]{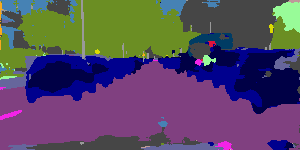}&\includegraphics[width=\fWidD]{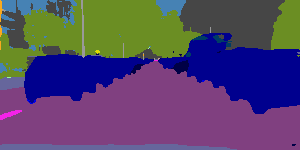}&\includegraphics[width=\fWidD]{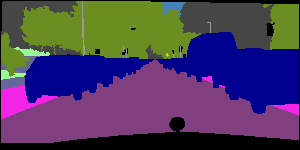}\\
\vspace{-0.07cm}\includegraphics[width=\fWidD]{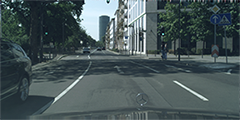}&\includegraphics[width=\fWidD]{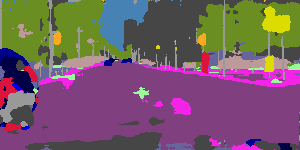}&\includegraphics[width=\fWidD]{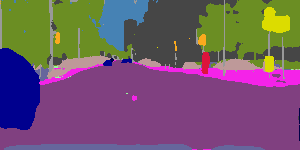}&\includegraphics[width=\fWidD]{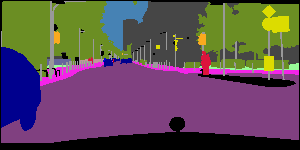}\\
\vspace{-0.07cm}\includegraphics[width=\fWidD]{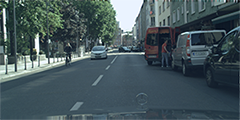}&\includegraphics[width=\fWidD]{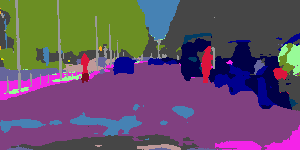}&\includegraphics[width=\fWidD]{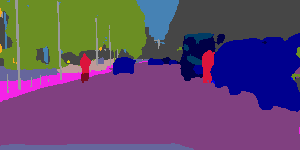}&\includegraphics[width=\fWidD]{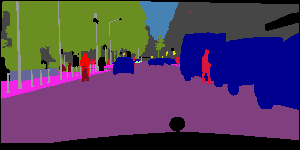}\\
\includegraphics[width=\fWidD]{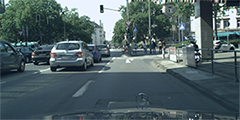}&\includegraphics[width=\fWidD]{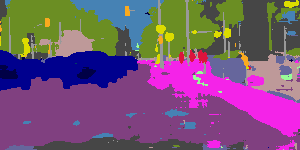}&\includegraphics[width=\fWidD]{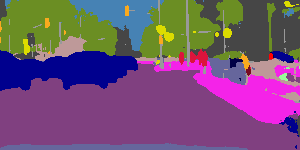}&\includegraphics[width=\fWidD]{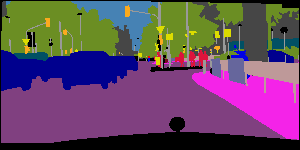}
\end{tabular}
}
\vspace{-0.1cm}
\caption{\sl\small {\bf The learned scene compatibility prior imposes regularity on the predictions.} When the scene compatibility is added, the segmentation network yields predictions better aligned with object boundaries, and are more consistent within the objects.}
\label{fig:cpn_result}
\vspace{-0.0cm}
\end{figure}

\begin{table*}[!ht]
\footnotesize \centering
\begin{tabular}{|c|c|p{0.23cm}<{\centering}p{0.23cm}<{\centering}p{0.23cm}<{\centering}p{0.23cm}<{\centering}p{0.23cm}<{\centering}p{0.23cm}<{\centering}p{0.23cm}<{\centering}p{0.23cm}<{\centering}p{0.23cm}<{\centering}p{0.23cm}<{\centering}p{0.23cm}<{\centering}p{0.23cm}<{\centering}p{0.23cm}<{\centering}p{0.23cm}<{\centering}p{0.23cm}<{\centering}p{0.23cm}<{\centering}p{0.1cm}<{\centering}p{0.23cm}<{\centering}p{0.4cm}<{\centering}|p{0.4cm}<{\centering}|p{0.4cm}<{\centering}|}
\hline
Method&\rot{Architecture}&\rot{road}&\rot{sidewalk}&\rot{building}&\rot{wall}&\rot{fence}&\rot{pole}&\rot{light}&\rot{sign}&\rot{vegetation}&\rot{terrain}&\rot{sky}&\rot{person}&\rot{rider}&\rot{car}&\rot{truck}&\rot{bus}&\rot{train}&\rot{motorcycle}&\rot{bicycle}&\rot{mIoU}&\rot{fwIoU}
\\
\hline
AdaptSegNet \cite{tsai2018learning}&A&86.5&25.9&79.8&22.1&20.0&23.6&33.1&21.8&81.8&25.9&75.9&57.3&26.2&76.3&29.8&32.1&\bf{7.2}&29.5&32.5&41.4&75.5\\
DCAN \cite{wu2018dcan}&A&85.0&30.8&81.3&25.8&21.2&22.2&25.4&26.6&83.4&36.7&76.2&58.9&24.9&80.7&29.5&42.9&2.5&26.9&11.6&41.7&76.2\\
CyCADA \cite{hoffman2017cycada}&A&88.3&40.9&81.4&26.9&19.7&31.3&31.8&31.9&81.6&22.3&77.1&56.3&25.1&80.8&33.4&38.6&0.0&24.6&35.5&43.6&77.9\\
SSF-DAN \cite{Du_2019_ICCV}&A&90.3&38.9&81.7&24.8&22.9&30.5&37.0&21.2&84.8&38.8&76.9&58.8&30.7&\bf{85.7}&30.6&38.1&5.9&28.3&36.9&45.4&79.6\\
BDL \cite{li2019bidirectional}&A&\bf{91.0}&44.7&84.2&34.6&27.6&30.2&36.0&36.0&85.0&\bf{43.6}&83.0&58.6&31.6&83.3&35.3&\bf{49.7}&3.3&28.8&35.6&48.5&81.1\\
Ours&A&\bf{91.0}&\bf{49.2}&\bf{85.6}&\bf{37.2}&\bf{29.7}&\bf{33.7}&\bf{38.1}&\bf{39.2}&\bf{85.4}&35.4&\bf{85.1}&\bf{61.1}&\bf{32.8}&84.1&\bf{45.6}&46.9&0.0&\bf{34.2}&\bf{44.5}&\bf{50.5}&\bf{82.0}\\
\hline
AdaptSegNet \cite{tsai2018learning}&B&87.3&29.8&78.6&21.1&18.2&22.5&21.5&11.0&79.7&29.6&71.3&46.8&6.5&80.1&23.0&26.9&0.0&10.6&0.3&35.0&74.9\\
CyCADA \cite{hoffman2017cycada}&B&85.2&37.2&76.5&21.8&15.0&23.8&22.9&21.5&80.5&31.3&60.7&50.5&9.0&76.9&17.1&28.2&\bf{4.5}&9.8&0.0&35.4&73.8\\
DCAN \cite{wu2018dcan}&B&82.3&26.7&77.4&23.7&20.5&20.4&30.3&15.9&80.9&25.4&69.5&52.6&11.1&79.6&24.9&21.2&1.3&17.0&6.7&36.2&72.9\\
SSF-DAN \cite{Du_2019_ICCV}&B&88.7&32.1&79.5&\bf{29.9}&22.0&23.8&21.7&10.7&80.8&29.8&72.5&49.5&16.1&82.1&23.2&18.1&3.5&24.4&8.1&37.7&76.3\\
BDL \cite{li2019bidirectional}&B&89.2&40.9&81.2&29.1&19.2&14.2&29.0&19.6&\bf{83.7}&35.9&\bf{80.7}&\bf{54.7}&23.3&\bf{82.7}&\bf{25.8}&28.0&2.3&\bf{25.7}&19.9&41.3&78.4\\
Ours&B&\bf{90.2}&\bf{44.7}&\bf{82.0}&28.4&\bf{28.4}&\bf{24.4}&\bf{33.7}&\bf{35.6}&\bf{83.7}&\bf{40.5}&75.1&54.4&\bf{28.2}&80.3&23.8&\bf{39.4}&0.0&22.8&\bf{30.8}&\bf{44.6}&\bf{79.3}\\
\hline
CyCADA \cite{hoffman2017cycada}&C&79.1&33.1&77.9&23.4&17.3&32.1&33.3&31.8&81.5&26.7&69.0&\bf{62.8}&\bf{14.7}&74.5&\bf{20.9}&25.6&6.9&\bf{18.8}&20.4&39.5&72.7\\
Ours&C&\bf{90.7}&\bf{49.8}&\bf{81.9}&\bf{23.4}&\bf{18.5}&\bf{37.3}&\bf{35.5}&\bf{34.3}&\bf{82.9}&\bf{36.5}&\bf{75.8}&61.8&12.4&\bf{83.2}&19.2&\bf{26.1}&4.0&14.3&\bf{21.8}&\bf{42.6}&\bf{79.7}\\
\hline
\end{tabular}
\vspace{0.15cm}
\caption{\sl\small {\bf Quantitative Evaluation on the GTA5-to-Cityscapes benchmark.} Our method achieves the best mIoU and fwIoU using different segmentation architectures: A (ResNet-101), B (VGG-16), C (DRN-26). }
\vspace{-0.1cm}
\label{tab:gta5_result}
\end{table*}

\begin{table*}[ht]
\footnotesize \centering
\begin{tabular}{|c|c|p{0.31cm}<{\centering}p{0.31cm}<{\centering}p{0.31cm}<{\centering}p{0.31cm}<{\centering}p{0.31cm}<{\centering}p{0.31cm}<{\centering}p{0.31cm}<{\centering}p{0.31cm}<{\centering}p{0.31cm}<{\centering}p{0.31cm}<{\centering}p{0.31cm}<{\centering}p{0.31cm}<{\centering}p{0.31cm}<{\centering}p{0.31cm}<{\centering}p{0.31cm}<{\centering}p{0.4cm}<{\centering}|c|c|}
\hline
Method&\rot{Architecture}&\rot{road}&\rot{sidewalk}&\rot{building}&\rot{wall*}&\rot{fence*}&\rot{pole*}&\rot{light}&\rot{sign}&\rot{vegetation}&\rot{sky}&\rot{person}&\rot{rider}&\rot{car}&\rot{bus}&\rot{motorcycle}&\rot{bicycle}&mIoU&mIoU*
\\
\hline
AdaptPatch \cite{tsai2019domain}&A&82.4&38.0&78.6&8.7&0.6&26.0&3.9&11.1&75.5&84.6&53.5&21.6&71.4&32.6&19.3&31.7&40.0&46.5\\
AdaptSegNet \cite{tsai2018learning}&A&84.3&42.7&77.5&-&-&-&4.7&7.0&77.9&82.5&54.3&21.0&72.3&32.2&18.9&32.3&-&46.7\\
SSF-DAN \cite{Du_2019_ICCV}&A&84.6&41.7&80.8&-&-&-&11.5&14.7&\bf{80.8}&\bf{85.3}&\bf{57.5}&21.6&\bf{82.0}&36.0&19.3&34.5&-&50.0\\
BDL \cite{li2019bidirectional}&A&\bf{86.0}&\bf{46.7}&80.3&-&-&-&14.1&11.6&79.2&81.3&54.1&27.9&73.7&\bf{42.2}&25.7&45.3&-&51.4\\
Ours&A&85.9&44.6&\bf{80.8}&\bf{9.0}&\bf{0.8}&\bf{32.1}&\bf{24.8}&\bf{23.1}&79.5&83.1&57.2&\bf{29.3}&73.5&34.8&\bf{32.4}&\bf{48.2}&\bf{46.2}&\bf{53.6}\\
\hline
AdaptSegNet \cite{tsai2018learning}&B&78.9&29.2&75.5&-&-&-&0.1&4.8&72.6&76.7&43.4&8.8&71.1&16.0&3.6&8.4&-&37.6\\
AdaptPatch \cite{tsai2019domain}&B&72.6&29.5&77.2&\bf{3.5}&0.4&21.0&1.4&7.9&73.3&79.0&45.7&14.5&69.4&19.6&7.4&16.5&33.7&39.6\\
DCAN \cite{wu2018dcan}&B&\bf{79.9}&30.4&70.8&1.6&\bf{0.6}&22.3&6.7&23.0&76.9&73.9&41.9&16.7&61.7&11.5&10.3&38.6&35.4&41.7\\
BDL \cite{li2019bidirectional}&B&72.0&30.3&74.5&0.1&0.3&\bf{24.6}&\bf{10.2}&25.2&\bf{80.5}&80.0&\bf{54.7}&23.2&\bf{72.7}&24.0&7.5&44.9&39.0&46.1\\
Ours&B&79.7&\bf{35.2}&\bf{78.7}&1.4&\bf{0.6}&23.1&10.0&\bf{28.9}&79.6&\bf{81.2}&51.2&\bf{25.1}&72.2&\bf{24.1}&\bf{16.7}&\bf{50.4}&\bf{41.1}&\bf{48.7}\\
\hline
\end{tabular}
\vspace{0.15cm}
\caption{\sl\small {\bf Quantitative Evaluation on the Synthia-to-Cityscapes Benchmark.} mIoU and mIoU* are the mean IoU computed on the 16 classes and the 13 subclasses respectively (* excluded). Our method achieves the best performance using different segmentation network backbones: A (ResNet-101), B(VGG-16). }
\vspace{-0.1cm}
\label{tab:synthia_result}
\end{table*}

\textbf{Phase Consistency:} Here we train the segmentation network Deeplab-V2 \cite{CP2016Deeplab} on the transformed source dataset with phase consistency. To make the comparison fair, all competing methods also use the same segmentation network as ours. The results of \cite{tsai2018learning} and \cite{li2019bidirectional} are reported by the original papers. We retrain \cite{hoffman2017cycada} and report its best performance with hyperparameter tuning. The result is presented in Tab. \ref{tab:phase_result}. Without any surrogate semantic consistency provided by a surrogate semantic segmentation network, our segmentation model achieves higher accuracy. Note that introducing surrogate semantic consistency for regularizing the transformation networks will also incur more memory cost. Moreover, several rounds of training to improve the performance of the surrogate network can also be time-consuming. However, our phase consistency can be implemented at low computational overhead (see Sect. \ref{sec:computational_cost}).

Interestingly, output space regularization, which aligns the marginal distributions of the segmentations, occasionally leads to worse performance in some settings, including \cite{hoffman2017cycada} and ours. This is somewhat reasonable since aligning the marginal distributions does not guarantee the conditional alignment given the observations.

\textbf{Scene Compatibility:}
To better understand the performance gain from the scene compatibility prior, we compare to competing methods on the same transformed source images. We collect the scores for all the other methods using the same setting as ours, if needed, we retrain their model.

In Tab. \ref{tab:cpn_result}, we show that under all experimental settings, scene compatibility prior improves accuracy for most of the semantic classes as well as the overall average. The performance gain is preserved during self-supervised learning. We present qualitative comparisons in Fig. \ref{fig:cpn_result}, showing that the scene compatibility prior provides strong spatial regularity to align segmentation to the object boundaries. Incorporating the scene compatibility prior into the training process significantly improves the overall segmentation smoothness and integrity, resulting in more consistent label prediction within each object. 
%Note that, one may also refine the segmentation output by CPN directly as post-processing for more scene compatible results, which typically leads to sharper segmentation boundaries and improves the mIoU by 1\%. 
%However, this practice requires loading CPN at time of inference, which leads to additional computational time and memory, thus is not applied to our experiments.

\subsection{Benchmark Results}

\def\figd{figures/gta}
\def\fWidD{0.142\textwidth}
\begin{figure*}[!t]
\hspace*{-5mm}\centering
{\footnotesize
\begin{tabular}{c@{\hspace{0.01in}}c@{\hspace{0.01in}}c@{\hspace{0.01in}}c@{\hspace{0.01in}}c@{\hspace{0.01in}}c@{\hspace{0.01in}}c@{\hspace{0.01in}}c}\\

\vspace{-0.15cm}\rot{\quad Image}&\includegraphics[width=\fWidD]{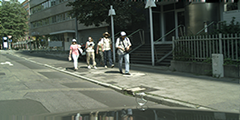}&\includegraphics[width=\fWidD]{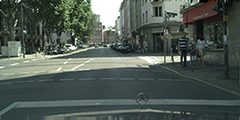}&\includegraphics[width=\fWidD]{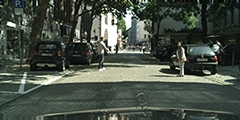}&\includegraphics[width=\fWidD]{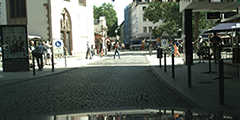}&\includegraphics[width=\fWidD]{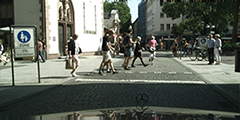}&\includegraphics[width=\fWidD]{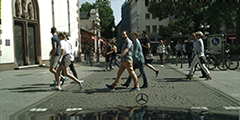}&\includegraphics[width=\fWidD]{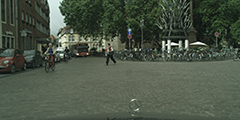}\\
\vspace{-0.07cm} \rot{\scriptsize \!\!\! CyCADA \cite{hoffman2017cycada}}&\includegraphics[width=\fWidD]{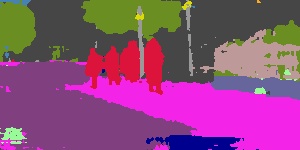}&\includegraphics[width=\fWidD]{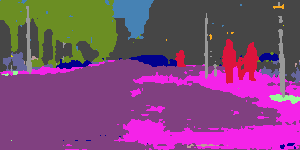}&\includegraphics[width=\fWidD]{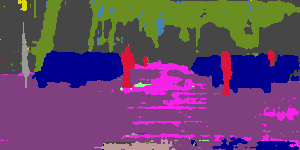}&\includegraphics[width=\fWidD]{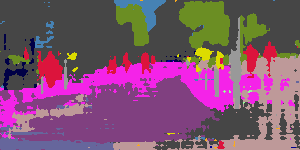}&\includegraphics[width=\fWidD]{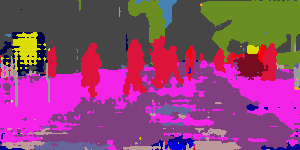}&\includegraphics[width=\fWidD]{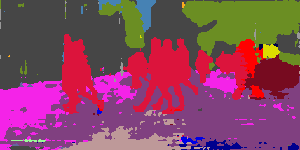}&\includegraphics[width=\fWidD]{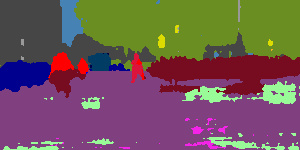}\\
\vspace{-0.07cm}\rot{BDL \cite{li2019bidirectional}}&\includegraphics[width=\fWidD]{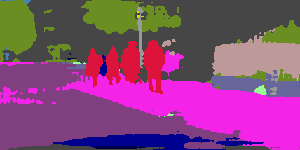}&\includegraphics[width=\fWidD]{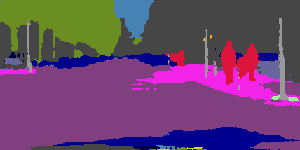}&\includegraphics[width=\fWidD]{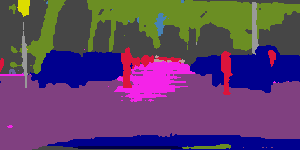}&\includegraphics[width=\fWidD]{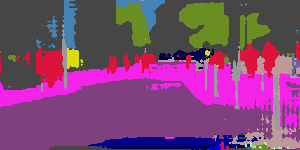}&\includegraphics[width=\fWidD]{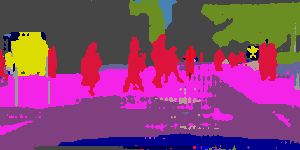}&\includegraphics[width=\fWidD]{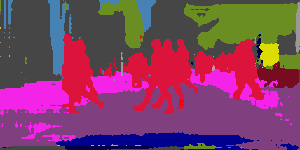}&\includegraphics[width=\fWidD]{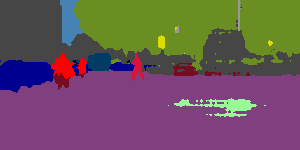}\\
\vspace{-0.07cm}\rot{\quad Phase}&\includegraphics[width=\fWidD]{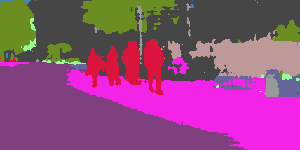}&\includegraphics[width=\fWidD]{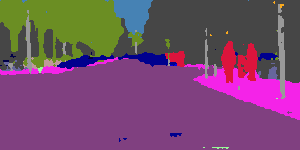}&\includegraphics[width=\fWidD]{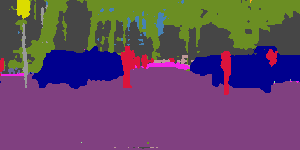}&\includegraphics[width=\fWidD]{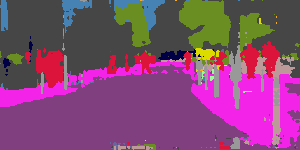}&\includegraphics[width=\fWidD]{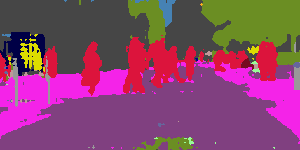}&\includegraphics[width=\fWidD]{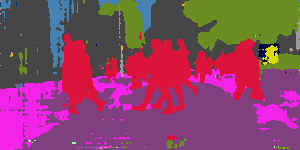}&\includegraphics[width=\fWidD]{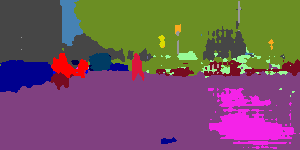}\\
\vspace{-0.07cm}\rot{\quad\,\,Full}&\includegraphics[width=\fWidD]{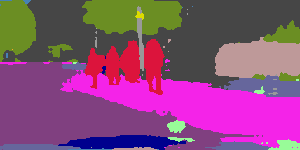}&\includegraphics[width=\fWidD]{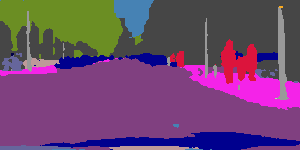}&\includegraphics[width=\fWidD]{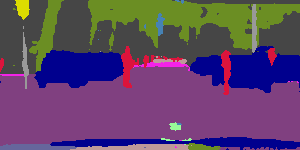}&\includegraphics[width=\fWidD]{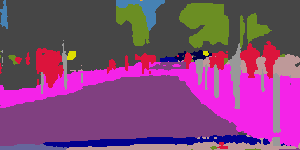}&\includegraphics[width=\fWidD]{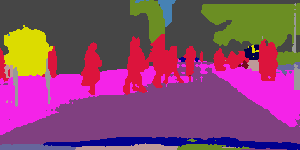}&\includegraphics[width=\fWidD]{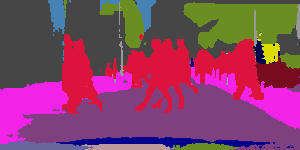}&\includegraphics[width=\fWidD]{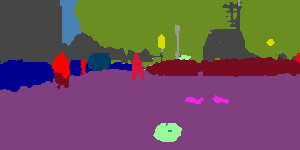}\\
\rot{\quad \,\,\,GT}&\includegraphics[width=\fWidD]{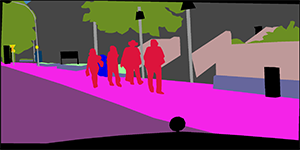}&\includegraphics[width=\fWidD]{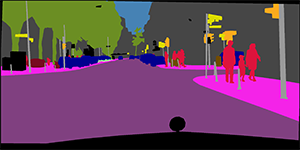}&\includegraphics[width=\fWidD]{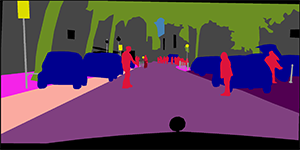}&\includegraphics[width=\fWidD]{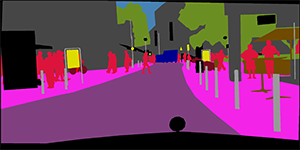}&\includegraphics[width=\fWidD]{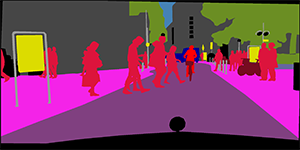}&\includegraphics[width=\fWidD]{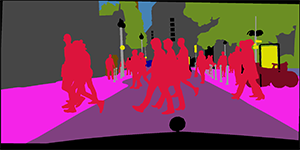}&\includegraphics[width=\fWidD]{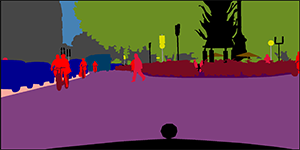}
\end{tabular}
}
\caption{\sl\small {\bf Qualitative comparison with state-of-the-art methods.} Our method outputs more spatially regularized segmentations align well with the underlying scene structure. All visualized models are based on DeepLab-V2 with ResNet-101 under the same setting. Phase: trained with phase consistency only; Full: our full model.}
\vspace{-0.3cm}
\label{fig:gta_result}
\end{figure*}

To recall, feature space alignment has been explored by \textbf{DCAN} \cite{wu2018dcan} and \textbf{CyCADA} \cite{hoffman2017cycada}. CyCADA also applies image level domain alignment by training cross-domain cycle consistent image transformation. Output space alignment methods include \textbf{AdaptSegNet} \cite{tsai2018learning}, \textbf{AdaptPatch} \cite{tsai2019domain} and \textbf{SSF-DAN} \cite{Du_2019_ICCV}, in which various ways of adversarial learning to the segmentation output are applied for better domain confusion. \textbf{BDL} \cite{li2019bidirectional} propagates information from semantic segmentation back to the image transformation network as semantic consistent regularization. 

We apply ResNet-101 \cite{He2015DeepRL} based Deeplab-V2 \cite{CP2016Deeplab} and VGG-16 \cite{Simonyan2014VeryDC} based FCN-8s \cite{Shelhamer2014FullyCN} for the segmentation network to compare with \cite{tsai2018learning,wu2018dcan,li2019bidirectional,vu2019advent,Du_2019_ICCV} under the same experimental setting. To better understand the robustness to different neural network settings, We also apply our method to the retrain the DRN-26 \cite{Yu2017} model from \cite{hoffman2017cycada}.

The result on the GTA5-to-Cityscapes benchmark is summarized in Tab. \ref{tab:gta5_result}. Our method achieves state-of-the-art performance with all network backbones in terms of mIoU and fwIoU. Moreover, across different settings, our method achieves the best score for most of the classes, indicating that the proposed priors improve the segmentation accuracy consistently across different semantic categories. We also present a qualitative comparison in Fig. \ref{fig:gta_result}. Our proposed method outputs more spatially regularized predictions, which are also consistent with the scene structures. We relatively achieve 4.1\% and 8.0\% improvement over the second-best method with the backbone ResNet-101 and VGG-16, respectively.

The result on the Synthia-to-Cityscapes benchmark can be found in Tab. \ref{tab:synthia_result}. The mIoUs of either 13 or 16 classes are evaluated according to the evaluation protocol in the literature. Our method outperforms competing methods on both sets. It also achieves the best result on most of the semantic categories. Again, we relatively achieve 4.3\% and 5.4\% improvement over the second-best using different backbones.

\subsection{Computational Cost} \label{sec:computational_cost}
All networks are trained using a single Nvidia Titan Xp GPU. Enforcing the phase consistency will incur a $<$0.001s overhead for a $1024 \times 512$ image, which is negligible. Training the CPN for scene compatibility takes 2.5 seconds to process a batch of 4 images, given the images are cropped to $1280 \times 768$. Incorporating CPN into segmentation training adds 1.5 seconds overhead to each iteration. Note that CPN is not required at the time of inference to segment target images.

\section{Discussion}
%\subsection{Limitations and Future Work}

It is empirically shown in Sect. \ref{sec:experiments} that the proposed priors improve UDA semantic segmentation accuracy under different settings, however, how to impose semantic consistency and ecological statistics priors to general UDA tasks besides semantic segmentation remains an open problem. 

Analysis of the CPN is another unsolved task. Currently, the capacity of the CPN bottleneck is chosen empirically. In order to estimate the optimal bottleneck capacity for specific tasks, quantitative measurement of the information that CPN leverages from the image is necessary, which requires future exploration.

%\subsection{Conclusion}
Unsupervised domain adaptation is key for semantic segmentation, where dense annotation in real images is costly and rare, but comes automatically in rendered images. UDA is a form of transfer learning that hinges on regularities and assumptions or priors on the relationship between the distributions from which the source and target data are sampled. We introduce two assumptions, and the corresponding priors and variational renditions that are integrated into end-to-end differential learning. One is that the transformations mapping one domain to another only affect the magnitude, but not the phase, of their spectrum. This is motivated by empirical evidence that image semantics, as perceived by the human visual system, go with the phase but not the magnitude of the spectrum. The other is a prior meant to capture the ecological statistics, that are characteristics of the images induced by regularities in the scene, and therefore shared across different imaging modalities and domains. We show that the resulting priors improve performance in UDA benchmarks, and quantify their impact through ablation studies.

\section*{Acknowledgements}

Research supported by ARO W911NF-17-1-0304 and ONR N00014-19-1-2066. Dong Lao is supported by KAUST through the VCC Center Competitive Funding.

{\small
\bibliographystyle{ieee_fullname}
\bibliography{main}
}

\end{document}